\documentclass{article}

\usepackage{arxiv}

\usepackage[utf8]{inputenc} % allow utf-8 input
\usepackage[T1]{fontenc}    % use 8-bit T1 fonts
\usepackage{hyperref}       % hyperlinks
\usepackage{url}            % simple URL typesetting
\usepackage{booktabs}       % professional-quality tables
\usepackage{amsfonts}       % blackboard math symbols
\usepackage{nicefrac}       % compact symbols for 1/2, etc.
\usepackage{microtype}      % microtypography
\usepackage{lipsum}

\usepackage{bm}
\usepackage{amsmath}
\usepackage{amssymb}
\usepackage{amsfonts}
\usepackage{graphicx}
\def\vec#1{\mathbf{#1}}
\usepackage{algorithm}
\usepackage{algorithmic}

\title{Few-shot Learning for Time-series Forecasting}

\author{
  Tomoharu Iwata\\
  NTT Communication Science Laboratories\\
  \And
  Atsutoshi Kumagai\\
  NTT Software Innovation Center\\
}
\date{}

\begin{document}
\maketitle

\begin{abstract}
Time-series forecasting is important for many applications. Forecasting models are usually trained using time-series data in a specific target task. However, sufficient data in the target task might be unavailable, which leads to performance degradation. In this paper, we propose a few-shot learning method that forecasts a future value of a time-series in a target task given a few time-series in the target task. Our model is trained using time-series data in multiple training tasks that are different from target tasks. Our model uses a few time-series to build a forecasting function based on a recurrent neural network with an attention mechanism. With the attention mechanism, we can retrieve useful patterns in a small number of time-series for the current situation. Our model is trained by minimizing an expected test error of forecasting next timestep values. We demonstrate the effectiveness of the proposed method using 90 time-series datasets.
\end{abstract}

% keywords can be removed

\section{Introduction}
Time-series forecasting is important for many applications,
which include
finantial markets~\cite{azoff1994neural,kim2003financial,cao2003support},
enegy management~\cite{deb2017review,sfetsos2000comparison},
traffic system~\cite{laptev2017time,jilani2007multivariate,li2017diffusion,yu2017spatio},
and
environmental engineering~\cite{lachtermacher1994backpropagation}.
Recently, deep learning methods,
such as Long Short Term Memory (LSTM)~\cite{hochreiter1997long},
have been widely used for time-series forecasting models
due to its high performance~\cite{assaad2008new,ogunmolu2016nonlinear,laptev2017time}.

Forecasting models are usually trained
using time-series data in a specific target task,
where we want to forecast future values.
For example, to train traffic congesting forecasting models,
we use traffic congestion time-series data at many locations.
However, sufficient data in the target task might be unavailable,
which leads to performance degradation.

\begin{figure}[t!]
  \centering
  \includegraphics[width=38em]{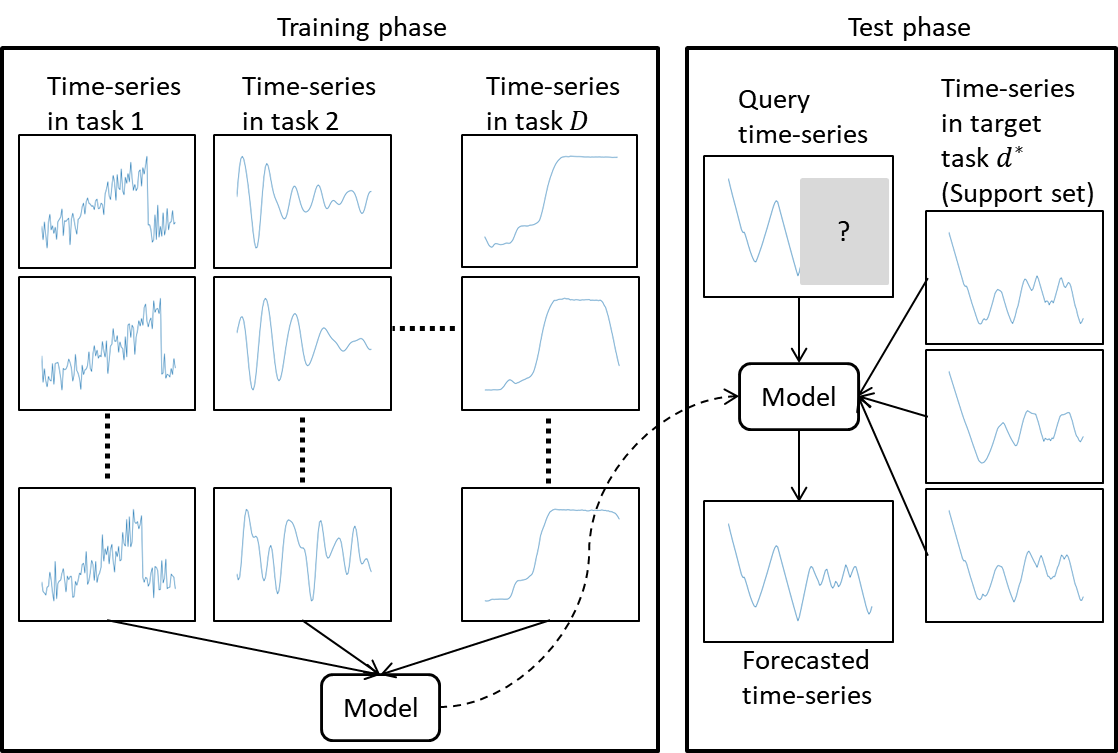}
  \caption{Our problem formulation. In a training phase, our model is trained by using time-series data in multiple tasks. In a test phase, given a few time-series data (support set) and query time-series, our model forecasts future values of the query time-series, where tasks given at the test phase are different from tasks given at the training phase.}
  \label{fig:task}  
\end{figure}

In this paper, we propose a few-shot learning method
that forecasts time-series in a target task given a few time-series,
where time-series in the target task are not given in a training phase.
The proposed method trains our model using time-series data
in multiple training tasks that are different from the target task.
Figure~\ref{fig:task} illustrates our problem formulation.
Time-series in other tasks might have similar dynamics
to those in the target task.
For example, many time-series include trend that shows the long-term tendency of the time-series to increase or decrease.
Also, time-series that are related to human activity, such as traffic volume and electric power consumption, exhibit daily and/or weekly cyclic dynamics.
By using knowledge learned from various time-series data, we can improve the forecasting performance of the target task.

Given a few time-series, which are called a support set,
our model outputs a value at the next timestep of a time-series,
which is called a query.
In particular, first, we obtain representations of the support set
with a bidirectional LSTM.
Then, we forecast future values of the query
considering the support representations
based on an attention mechanism
as well as the query's own pattern based on an LSTM.
With the attention mechanism,
we can retrieve useful patterns in the support sets
to forecast at the current situation.
In addition, we can handle the support set with the different number of
time-series with different length using the attention mechanism.
Given a target task, our model forecasts future values
that are tailored to the target task without retraining.
Our model is trained by minimizing an expected test error
of forecasting next timestep values
given a support set, which are calculated using data in multiple training tasks.

The main contributions of this paper are:
\begin{enumerate}
\item Our method is the first method of few-shot learning for time-series forecasting that does not require retraining given target tasks.
\item Our model can handle different support size and different time-series length with an attention mechanism and LSTMs.
\item We demonstrate the effectiveness of the proposed method using 90 time-series datasets.
\end{enumerate}
The remainder of this paper is organized as follows.
In Section~\ref{sec:related}, we briefly review related work.
In Section~\ref{sec:proposed},
we propose our model and its training procedure for few-shot time-series forecasting.
In Section~\ref{sec:experiments},
we show that the proposed method outperforms existing methods.
Finally, we give a concluding remark and future work
in Section~\ref{sec:conclusion}.

\section{Related work}
\label{sec:related}

For transfer knowledge in source tasks to target tasks,
many transfer learning, domain adaptation,
and multi-task learning methods have been
proposed~\cite{tan2018survey,long2017deep,killian2017robust,jia2018transfer,kumagai2019transfer}.
%There are some applications of transfer learning to time-series forecasting~\cite{gupta2018transfer,xiao2014transfer,qureshi2017wind,hu2016transfer}.
However, these methods require relatively a large number
of time-series of target tasks.
To reduce the required number of target examples,
few-shot learning, or meta-learning, has been attract considerable attention
recently~\cite{schmidhuber:1987:srl,bengio1991learning,ravi2016optimization,andrychowicz2016learning,vinyals2016matching,snell2017prototypical,bartunov2018few,finn2017model,li2017meta,kimbayesian,finn2018probabilistic,rusu2018meta,yao2019hierarchically,edwards2016towards,garnelo2018conditional,kim2019attentive,hewitt2018variational,bornschein2017variational,reed2017few,rezende2016one,tang2019,narwariya2020meta,xie2019meta,lake2019compositional}.
There are some applications of few-shot learning to time-series forecasting~\cite{hooshmand2019energy,ribeiro2018transfer,lemke2010meta,prudencio2004meta,talagala2018meta,ali2018cross}.
Existing few-shot time-series forecasting methods can be categorized into two: finetune-based and meta feature-based.
Finetune-based methods~\cite{hooshmand2019energy,ribeiro2018transfer}
train models using training tasks, and finetune the models given target tasks.
On the other hand, the proposed method does not need to retrain the model
given target tasks.
Meta feature-based methods~\cite{lemke2010meta,prudencio2004meta,talagala2018meta,ali2018cross}
use meta features of time-series, such as standard deviation and length, to select forecasting models.
In contrast, the proposed method does not require to determine meta features;
it extracts latent representations of time-series with LSTMs.
Neural network-based time-series forecasting models~\cite{oreshkin2019n},
which can be considered as a meta-learning method,
are used for zero-shot time-series forecasting~\cite{oreshkin2020meta}.
Since they consider zero-shot learning,
where no target examples are given,
they cannot use given target time-series data.
Recurrent attentive neural processes~\cite{qin2019recurrent} uses recurrent neural networks with an attention mechanism
for meta-learning, where attentions are connected to past sequences
for extending neural processes~\cite{kim2019attentive,willi2019recurrent,garnelo2018conditional}.
Therefore, they cannot use different time-series data given as a support set.
On the other hand, the proposed method connects attentions to time-series data in a support set
to use them for improving the performance.

\begin{figure}
  \centering
  \includegraphics[width=20em]{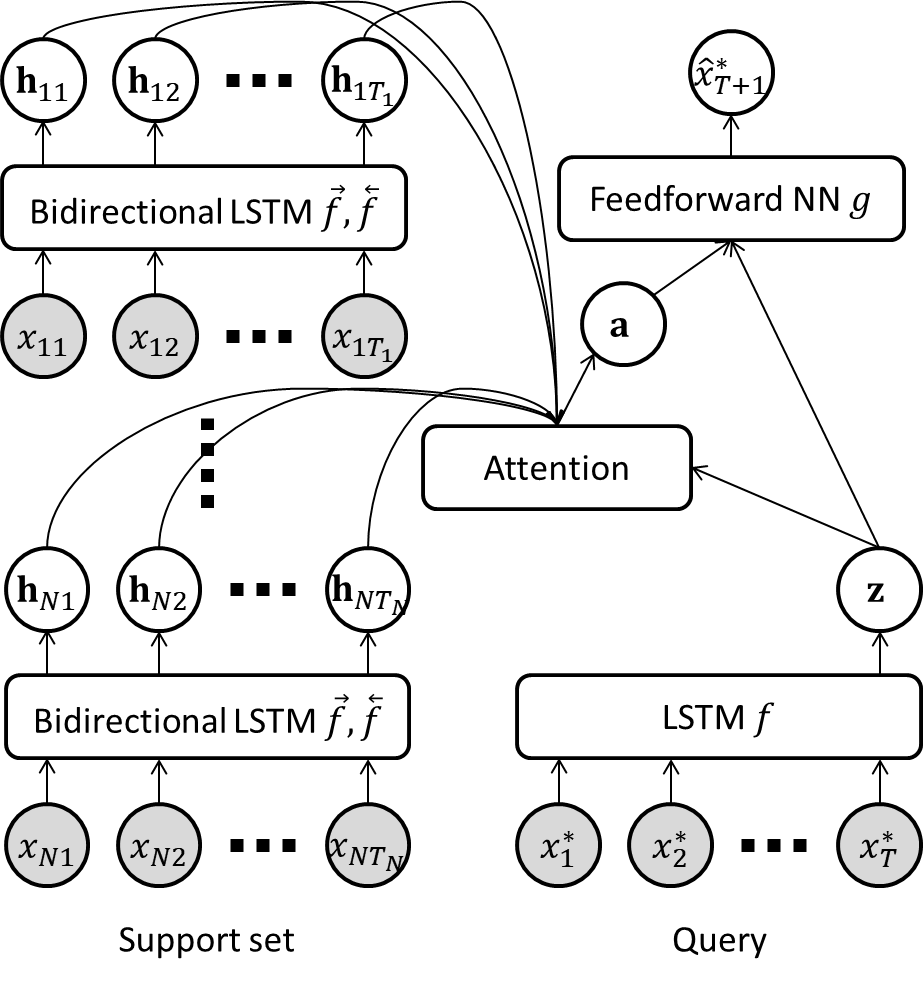}
  \caption{
    Our model. Time-series $\{\{x_{nt}\}_{t=1}^{T_{n}}\}_{n=1}^{N}$ in a support set are transformed into hidden states $\{\{\vec{h}_{nt}\}_{t=1}^{T_{n}}\}_{n=1}^{N}$ by a bidirectional LSTM.
    Query time-series $\{x_{t}^{*}\}_{t=1}^{T}$ is transformed into hidden state $\vec{z}$ by LSTM $f$. An attention mechanism outputs $\vec{a}$ by giving attention to support hidden states $\{\{\vec{h}_{nt}\}_{t=1}^{T_{n}}\}_{n=1}^{N}$
    based on query hidden state $\vec{z}$. Our model predicts value at the next timestep $\hat{x}^{*}_{T+1}$
    by feeding attention mechanism output $\vec{a}$ and query hidden state $\vec{z}$ to feed-forward neural network $g$.}
  \label{fig:model}
\end{figure}

\section{Proposed method}
\label{sec:proposed}

We describe our model that uses a support set to build a forecasting function
in Section~\ref{sec:model}.
In Section~\ref{sec:train}, we present the training procedure for our model given sets of time-series
in multiple tasks.
Then, we describe a test phase in Section~\ref{sec:test}.

\subsection{Model}
\label{sec:model}

Let $\mathcal{S}=\{\vec{x}_{n}\}_{n=1}^{N}$ be a support set,
where $\vec{x}_{n}=[x_{n1},\dots,x_{nT_{n}}]$ is the $n$th time-series,
$x_{nt}\in\mathbb{R}$ is a scalar continuous value at timestep $t$,
$T_{n}$ is its length, and $N$ is the number of time-series in the support set.
Our model uses support set $\mathcal{S}$ to build a forecasting function that
outputs predictive value $\hat{x}_{T+1}^{*}$ at the next timestep given query time-series $\vec{x}^{*}=[x_{1}^{*},\ldots,x_{T}^{*}]$
in the same task with the support set.
Figure~\ref{fig:model} illustrates our model.

First, we obtain representations of each timestep for each time-series in suppot set $\mathcal{S}$
using a bidirectional LSTM in the form of hidden states:
\begin{align}
  \overrightarrow{\vec{h}}_{nt}=\overrightarrow{f}(\overrightarrow{\vec{h}}_{n,t-1},x_{nt}),\nonumber\\
  \overleftarrow{\vec{h}}_{nt}=\overleftarrow{f}(\overleftarrow{\vec{h}}_{n,t+1},x_{nt}),
\end{align}
where $\overrightarrow{f}$ and $\overleftarrow{f}$ are forward and backword LSTMs,
and $\overrightarrow{\vec{h}}_{nt}\in\mathbb{R}^{K_{\overrightarrow{\mathrm{h}}}}$
and $\overleftarrow{\vec{h}}_{nt}\in\mathbb{R}^{K_{\overleftarrow{\mathrm{h}}}}$
are foward and backword hidden states
of the $n$th support time-series at timestep $t$.
The forward (backword) hidden state $\overrightarrow{\vec{h}}_{nt}$ ($\overleftarrow{\vec{h}}_{nt}$) contains
information about the time-series before (after) timestep $t$.
We use concatenated vector of the forward and backward hidden states,
$\vec{h}_{nt}=[\overrightarrow{\vec{h}}_{nt},\overleftarrow{\vec{h}}_{nt}]\in\mathbb{R}^{K_{\mathrm{h}}}$,
as the representation of the $n$th time-series at timestep $t$ 
where 
$[\cdot,\cdot]$ represents the concatenation of vectors, and
$K_{\mathrm{h}}=K_{\overrightarrow{\mathrm{h}}}+K_{\overleftarrow{\mathrm{h}}}$.
With the bidirectional LSTM, we can encode both past and future information in representation $\vec{h}_{nt}$,
which is important for forecasting.
In addition, LSTMs enable us to handle time-series in different lengths.

Second, we obtain a representation of query time-series $\vec{x}^{*}$ with LSTM $f$:
\begin{align}
  \vec{z}_{t}=f(\vec{z}_{t-1},x_{t}^{*}),
\end{align}
where $\vec{z}_{t}\in\mathbb{R}^{K_{\mathrm{z}}}$ is the hidden state at timestep $t$.
We use the hidden state at the last timestep $T$ as query's representation $\vec{z}:=\vec{z}_{T}$.

Third, we extract knowledge from the support set
that is useful for forecasting using an attention mechanism:
\begin{align}
  %\mathrm{softmax}((\vec{K}\vec{\vec{h}})^{\top}\vec{Q}\vec{z})\vec{V}\vec{\vec{h}}
  \vec{a}=\sum_{n=1}^{N}\sum_{t=1}^{T_{n}}\frac{\exp\left((\vec{K}\vec{h}_{nt})^{\top}\vec{Q}\vec{z}\right)}
      {\sum_{n'=1}^{N}\sum_{t'=1}^{T_{n'}}\exp\left((\vec{Q}\vec{z})^{\top}\vec{K}\vec{h}_{n't'}
        \right)}\vec{V}\vec{h}_{nt},
\end{align}
where  
$\vec{Q}\in\mathbb{R}^{K_{\mathrm{a}}\times K_{\mathrm{z}}}$,
$\vec{K}\in\mathbb{R}^{K_{\mathrm{a}}\times K_{\mathrm{h}}}$
and 
$\vec{V}\in\mathbb{R}^{K_{\mathrm{v}}\times K_{\mathrm{h}}}$
are linear projection matrices.
When there are support time-series that have locally similar patterns with the query,
the attention mechanism retrieves information of the point, $\vec{V}\vec{h}_{nt}$.
The similarity is calculated by the inner product between linearly transformed support representations, $\vec{K}\vec{h}_{nt}$ and linearly transformed query representations, $\vec{Q}\vec{z}$.
By training our model so as to
minimize the expected forecasting error as described in Section~\ref{sec:train},
the attention mechanism retrieves information that is effective to
improve the forecasting performance.
Since parameters of the attention mechanism, $\vec{Q}$, $\vec{K}$ and $\vec{V}$,
do not depend on the number of time-series in the support set,
we can deal with support sets with different sizes.

Then,
we forecast a value at next timestep $t+1$ using
both attention output $\vec{a}$ and query representation $\vec{z}$:
\begin{align}
  %\hat{x}^{*}_{T+1}=
  \mu(\vec{x}^{*};\mathcal{S},\bm{\Phi}):=g([\vec{a},\vec{z}]),
\end{align}
where $g$ is a feed-forward neural network,
and $\bm{\Phi}$ is parameters in our model,
which are parameters of bidirectional LSTMs, 
$\overrightarrow{f}$, $\overleftarrow{f}$, LSTM $f$, and
feed-forward neural network $g$,
linear projection matrices in the attention mechanism,
$\vec{Q}$, $\vec{K}$ and $\vec{V}$.
By including query representation $\vec{z}$ in the input of the
neural network,
we can forecast using its own past values $\vec{x}^{*}$
even if there is no useful information in the support set.

\subsection{Training}
\label{sec:train}

In a training phase, we are given a set of one-dimensional time-series in $|\mathcal{D}|$ tasks
$\mathcal{X}=\{\vec{X}_{d}\}_{d\in\mathcal{D}}$,
where $\vec{X}_{d}=\{\vec{x}_{dn}\}_{n=1}^{N_{d}}$ is the set of time-seres in task $d$,
$\vec{x}_{dn}=[x_{dn1},\dots,x_{dnT_{dn}}]$ is the $n$th time-series in task $d$,
$x_{dnt}\in\mathbb{R}$ is a scalar continuous value at timestep $t$,
$T_{dn}$ is its length, and $N_{d}$ is the number of time-series in task $d$.

We estimate model parameters $\bm{\Phi}$ 
by minimizing the expected loss on a query set given a support set
using an episodic training framework,
where support and query sets are randomly generated from
training datasets $\mathcal{X}$
to simulate target tasks:
\begin{align}
  \hat{\bm{\Phi}}=
  \arg \min_{\bm{\Phi}} \mathbb{E}_{d\sim\mathcal{D}}[
    \mathbb{E}_{(\mathcal{S},\mathcal{Q})\sim\vec{X}_{d}}[
      L(\mathcal{S},\mathcal{Q};\bm{\Phi})]],
\end{align}
where $\mathbb{E}$ represents an expectation,
\begin{align}
  L(\mathcal{S},\mathcal{Q};\bm{\Phi})
  =  \frac{1}{N_{\mathrm{Q}}}
  \sum_{n=1}^{N_{\mathrm{Q}}}
  \frac{1}{T_{n}}\sum_{t=1}^{T_{n}}
  \parallel \mu(\vec{x}_{n,:t-1},\mathcal{S};\bm{\Phi})-x_{nt}\parallel^{2},
  \label{eq:E}
\end{align}
is the mean squared error on the predictions at the next timestep values in query set $\mathcal{Q}$ given support set $\mathcal{S}$,
$N_{\mathrm{Q}}$ is the number of instances in the query set,
$T_{n}$ is length of the $n$th time-series in the query set,
$x_{nt}$ is the value of the $n$th sequence at timestep $t$,
and $\vec{x}_{n,:t-1}=[x_{n1},\ldots,x_{n,t-1}]$ is the time-series until timestep $t-1$.

The training procedure of our model is shown in
Algorithm~\ref{alg}.
For each iteration,
we randomly generate support and query sets (Lines 3 -- 5) from a randomly selected task.
Given the support and query sets, we calculate the loss (Line 6) by (\ref{eq:E}).
We update model parameters by using stochastic gradient descent methods (Line 7).

%In (\ref{eq:prediction}),
%mean function $f_{\mathrm{m}}([\vec{x},\vec{z}])$
%can be seen as a predictin before observing the support set,
%and the second term $\vec{k}^{\top}\vec{K}^{-1}(\vec{y}-\vec{m})$
%can be seen as a fine-tuning to the support set.

\begin{algorithm}[t!]
  \caption{Training procedure of our model.}
  \label{alg}
  \begin{algorithmic}[1]
    \REQUIRE{Time-series datasets $\mathcal{X}$,
      support set size $N_{\mathrm{S}}$,
      query set size $N_{\mathrm{Q}}$}
    \ENSURE{Trained model parameters $\bm{\Phi}$}
    \STATE Initialize parameters $\bm{\Phi}$ randomly
    \WHILE{not done}
    \STATE Sample task $d$ from $\mathcal{D}$
    \STATE Sample support set $\mathcal{S}$ with size $N_{\mathrm{S}}$ from $\{\vec{x}_{dn}\}_{n=1}^{N_{d}}$
    \STATE Sample query set $\mathcal{Q}$ with size $N_{\mathrm{Q}}$ from $\{\vec{x}_{dn}\}_{n=1}^{N_{d}}\setminus\mathcal{S}$    
    \STATE Calculate loss $L(\mathcal{S},\mathcal{Q};\bm{\Phi})$ by Eq.~(\ref{eq:E}) and its gradients 
    \STATE Update parameters $\bm{\Phi}$ using the gradients
    \ENDWHILE
  \end{algorithmic}
\end{algorithm}

\subsection{Test}
\label{sec:test}

In a test phase, we are given a few time-series in a new task $d^{*}\notin\mathcal{D}$ as a support set.
Then, we obtain a model that forecasts value $x_{d^{*},T^{*}+1}$ at the next timestep given
query time-series $\vec{x}_{d^{*}}=[x_{d^{*}1},\ldots,x_{d^{*}T}]$ in task $d^{*}$.

\section{Experiments}
\label{sec:experiments}

\subsection{Data}

We evaluated the proposed method 
using time-series datasets obtained from
UCR Time Series Classification Archive~\cite{UCRArchive2018,dau2019ucr}.
There were originally time-series data in 128 tasks.
We omit tasks that contain missing values,
time-series with the length shorter than 100,
and less than 50 time-series.
Then, we obtained time-series data in 90 tasks.
We used values at first 100 timesteps for each time-series.
We randomly split into 55 training, 10 validation and 25 target tasks,
where each task contains 50 time-series.
We normalized the values for each task with mean zero and variance one.

\subsection{Our model setting}

We used bidirectional LSTM $\overrightarrow{f}, \overleftarrow{f}$ with $K_{\overrightarrow{h}}=K_{\overleftarrow{h}}=32$ hidden units
for encoding support sets,
and LSTM $f$ with $K_{\mathrm{Z}}=32$ hidden units for encoding query sets.
In the attention mechanism, we used $K_{\mathrm{a}}=32$ and $K_{\mathrm{v}}=32$.
For the neural network to output a focasting value $g$,
we used three-layered feed-forward neural network with 32 hidden units.
The activation function in the neural networks were rectified linear unit, $\mathrm{ReLU}(x)=\max(0,x)$.
Optimization was performed using Adam~\cite{kingma2014adam} with learning rate $10^{-3}$ and
dropout rate $0.1$.
The maximum number of training epochs was 500,
and the validation datasets were used for early stopping.
We set support set size at $N_{S}=3$,
and query set size at $N_{Q}=47$.

\subsection{Comparing methods}

We compared the proposed method
with three types of training frameworks: model-agnostic meta-learning (MAML),
domain-independent learning (DI), and
domain-specific learning (DS).
With MAML,
initial model parameters are optimized so that they perform well when finetuned with a support set.
For the finetuning,
Adam with learning rate $10^{-3}$ and five epochs were used.
With DI, a model was trained by minimizing the error on all training tasks.
With DS, a model was trained by minimizing the error on the support set of the target task.
For MAML, DI, and DS, we used three types of models: LSTM, neural network (NN), and linear models (Linear).
With LSTM, we used LSTM with 32 hidden units. For forecasting values at the next timestep,
we used a three-layered feed-forward neural network with 32 hidden units
that takes the output of the LSTM.
With NN, we used three-layered feed-forward neural networks with 32 hidden units
that take values at one timestep before.
With Linear, we used linear regression models
that takes values at one timestep before.
We also compared with a method that output values that are the same with the previous timestep (Pre).

\subsection{Results}

\begin{table}[t!]
  \centering
  \begin{scriptsize}
  {\tabcolsep=0.7em
    \begin{tabular}{l|r|rrr|rrr|rrr|r} \hline
& & \multicolumn{3}{|c|}{LSTM} & \multicolumn{3}{|c|}{NN} & \multicolumn{3}{|c|}{Linear} & \\
      & Ours & MAML & DI & DS & MAML & DI & DS & MAML & DI & DS & Pre \\
      \hline
ACSF1 & {\bf 1.007} & {\bf 1.006} & {\bf 1.016} & 1.035 & 1.175 & 1.279 & 1.066 & 1.309 & 1.364 & 1.037 & 1.556\\
Adiac & {\bf 0.034} & 0.041 & {\bf 0.034} & 0.063 & 0.064 & 0.063 & 0.110 & 0.109 & 0.122 & 0.197 & 0.076\\
ArrowHead & {\bf 0.048} & {\bf 0.055} & {\bf 0.049} & 0.060 & 0.073 & 0.072 & 0.105 & 0.104 & 0.110 & 0.177 & 0.075\\
BME & {\bf 0.087} & {\bf 0.093} & {\bf 0.088} & 0.137 & 0.119 & 0.106 & 0.145 & 0.170 & 0.170 & 0.277 & 0.119\\
Beef & 0.047 & 0.070 & {\bf 0.039} & 0.080 & 0.070 & 0.067 & 0.121 & 0.108 & 0.119 & 0.371 & 0.073\\
CBF & 0.607 & 0.582 & 0.622 & {\bf 0.562} & 0.664 & 0.675 & 0.571 & 0.585 & 0.588 & 0.597 & 0.661\\
Car & {\bf 0.025} & 0.039 & {\bf 0.028} & 0.050 & 0.036 & 0.046 & 0.059 & 0.046 & 0.085 & 0.152 & 0.038\\
ChlorineConcentration & 0.551 & {\bf 0.497} & 0.571 & 0.513 & 0.520 & 0.554 & 0.533 & 0.529 & 0.562 & 0.553 & 0.556\\
CinCECGTorso & 0.152 & 0.152 & 0.154 & 0.231 & 0.197 & 0.156 & 0.175 & 0.177 & 0.166 & 0.284 & {\bf 0.138}\\
Coffee & {\bf 0.062} & {\bf 0.064} & 0.070 & 0.077 & 0.084 & 0.091 & 0.125 & 0.111 & 0.129 & 0.193 & 0.075\\
Computers & {\bf 0.398} & 0.515 & 0.407 & 0.751 & 0.663 & 0.451 & 0.596 & 0.473 & 0.477 & 0.544 & 0.553\\
CricketX & {\bf 0.395} & 0.423 & 0.407 & 0.459 & 0.439 & 0.424 & 0.456 & 0.486 & 0.459 & 0.492 & 0.455\\
CricketY & {\bf 0.400} & {\bf 0.394} & {\bf 0.392} & 0.412 & 0.405 & 0.431 & 0.422 & 0.453 & 0.431 & 0.479 & 0.441\\
CricketZ & {\bf 0.443} & 0.469 & 0.457 & 0.508 & 0.488 & 0.471 & 0.494 & 0.532 & 0.506 & 0.542 & 0.514\\
DiatomSizeReduction & {\bf 0.024} & {\bf 0.033} & {\bf 0.028} & 0.039 & 0.038 & 0.046 & 0.048 & 0.046 & 0.068 & 0.143 & 0.039\\
ECG5000 & {\bf 0.197} & {\bf 0.192} & 0.222 & {\bf 0.180} & 0.279 & 0.231 & 0.242 & 0.335 & 0.309 & 0.376 & 0.216\\
ECGFiveDays & {\bf 0.387} & {\bf 0.372} & 0.483 & {\bf 0.414} & 0.582 & 0.579 & 0.452 & 0.615 & 0.647 & 0.624 & 0.547\\
EOGHorizontalSignal & 0.160 & 0.160 & 0.159 & {\bf 0.164} & 0.161 & 0.162 & 0.183 & 0.185 & 0.180 & 0.236 & {\bf 0.153}\\
EOGVerticalSignal & 0.145 & 0.143 & 0.155 & 0.140 & 0.146 & 0.147 & 0.145 & 0.159 & 0.151 & 0.194 & {\bf 0.139}\\
Earthquakes & 1.054 & 1.016 & 1.087 & 1.002 & 1.033 & 1.108 & 1.007 & 1.168 & 1.215 & {\bf 0.997} & 1.397\\
EthanolLevel & 0.033 & {\bf 0.076} & 0.036 & 0.187 & 0.053 & 0.055 & {\bf 0.082} & 0.041 & 0.071 & 0.251 & {\bf 0.024}\\
FaceAll & {\bf 0.406} & {\bf 0.413} & 0.423 & 0.450 & 0.649 & 0.630 & 0.576 & 0.671 & 0.752 & 0.625 & 0.638\\
FaceFour & 0.357 & {\bf 0.333} & 0.353 & 0.363 & {\bf 0.337} & 0.361 & {\bf 0.358} & 0.378 & 0.393 & 0.456 & 0.346\\
FacesUCR & 0.442 & {\bf 0.401} & 0.458 & 0.516 & 0.802 & 0.661 & 0.548 & 0.710 & 0.779 & 0.631 & 0.647\\
FiftyWords & {\bf 0.051} & {\bf 0.059} & {\bf 0.056} & 0.181 & 0.095 & 0.101 & 0.164 & 0.167 & 0.179 & 0.258 & 0.120\\
Fish & {\bf 0.021} & 0.032 & 0.026 & 0.059 & 0.040 & 0.038 & 0.055 & 0.042 & 0.077 & 0.144 & 0.036\\
FordA & {\bf 0.107} & {\bf 0.111} & 0.135 & 0.200 & 0.273 & 0.292 & 0.336 & 0.429 & 0.420 & 0.511 & 0.316\\
FordB & {\bf 0.102} & {\bf 0.109} & 0.130 & 0.160 & 0.293 & 0.277 & 0.315 & 0.448 & 0.436 & 0.521 & 0.326\\
FreezerRegularTrain & {\bf 0.224} & 0.227 & 0.226 & 0.236 & {\bf 0.229} & 0.260 & 0.247 & 0.240 & 0.245 & 0.295 & {\bf 0.224}\\
FreezerSmallTrain & {\bf 0.227} & 0.227 & {\bf 0.226} & 0.253 & 0.228 & 0.255 & 0.240 & 0.237 & 0.238 & 0.292 & {\bf 0.224}\\
Fungi & {\bf 0.121} & {\bf 0.150} & {\bf 0.139} & 0.535 & 0.191 & 0.218 & 0.285 & 0.246 & 0.249 & 0.321 & 0.176\\
GunPoint & {\bf 0.064} & 0.072 & {\bf 0.065} & 0.123 & 0.072 & {\bf 0.071} & 0.109 & 0.109 & 0.122 & 0.191 & 0.088\\
GunPointAgeSpan & {\bf 0.067} & {\bf 0.070} & {\bf 0.067} & {\bf 0.068} & 0.071 & 0.085 & 0.099 & 0.083 & 0.106 & 0.177 & {\bf 0.067}\\
GunPointMaleVersusFemale & {\bf 0.056} & {\bf 0.064} & {\bf 0.055} & 0.067 & 0.118 & {\bf 0.065} & 0.099 & 0.090 & 0.102 & 0.244 & {\bf 0.056}\\
GunPointOldVersusYoung & {\bf 0.043} & 0.049 & {\bf 0.047} & 0.061 & 0.067 & 0.073 & 0.116 & 0.115 & 0.124 & 0.203 & 0.074\\
Ham & {\bf 0.127} & {\bf 0.123} & {\bf 0.131} & 0.301 & 0.255 & 0.226 & 0.302 & 0.326 & 0.311 & 0.426 & 0.221\\
HandOutlines & {\bf 0.031} & 0.052 & {\bf 0.032} & 0.101 & 0.046 & 0.058 & 0.101 & 0.059 & 0.104 & 0.224 & 0.041\\
Haptics & 0.372 & {\bf 0.307} & 0.373 & 0.404 & 0.478 & 0.412 & {\bf 0.378} & 0.486 & 0.497 & 0.554 & 0.363\\
Herring & {\bf 0.023} & 0.031 & 0.030 & 0.037 & 0.044 & 0.046 & 0.074 & 0.062 & 0.077 & 0.148 & 0.046\\
HouseTwenty & 0.388 & 0.392 & 0.401 & 0.383 & 0.387 & 0.400 & 0.379 & 0.404 & {\bf 0.369} & 0.416 & 0.389\\
InlineSkate & 0.058 & 0.066 & 0.064 & 0.092 & 0.075 & 0.073 & 0.082 & 0.055 & 0.087 & 0.182 & {\bf 0.043}\\
InsectEPGRegularTrain & 0.032 & 0.246 & 0.018 & 0.814 & 0.167 & 0.033 & 0.510 & 0.061 & 0.071 & 0.621 & {\bf 0.004}\\
InsectEPGSmallTrain & 0.030 & 0.093 & 0.022 & 0.685 & 0.516 & 0.036 & 0.484 & 0.025 & 0.064 & 0.545 & {\bf 0.002}\\
InsectWingbeatSound & {\bf 0.090} & 0.101 & 0.103 & 0.159 & 0.146 & 0.185 & 0.191 & 0.286 & 0.262 & 0.354 & 0.196\\
LargeKitchenAppliances & {\bf 0.846} & 0.919 & {\bf 0.876} & {\bf 0.862} & 1.176 & 1.212 & 0.955 & 1.133 & 1.114 & 1.040 & 1.325\\
Lightning2 & 0.042 & 0.056 & 0.044 & 0.079 & 0.054 & 0.055 & 0.059 & 0.064 & 0.084 & 0.185 & {\bf 0.026}\\
Lightning7 & {\bf 0.255} & {\bf 0.259} & 0.264 & 0.290 & 0.276 & 0.268 & 0.260 & {\bf 0.247} & 0.252 & 0.290 & 0.258\\
Mallat & {\bf 0.016} & 0.033 & 0.023 & 0.040 & 0.046 & 0.050 & 0.139 & 0.061 & 0.110 & 0.174 & 0.038\\
Meat & {\bf 0.049} & {\bf 0.055} & 0.054 & 0.167 & 0.093 & 0.084 & 0.151 & 0.178 & 0.182 & 0.276 & 0.122\\
MedicalImages & {\bf 0.173} & {\bf 0.183} & 0.209 & 0.327 & 0.250 & 0.248 & 0.363 & 0.370 & 0.351 & 0.403 & 0.251\\
MixedShapesRegularTrain & {\bf 0.047} & {\bf 0.051} & {\bf 0.047} & 0.081 & 0.068 & 0.064 & 0.070 & 0.098 & 0.091 & 0.188 & 0.051\\
MixedShapesSmallTrain & {\bf 0.031} & {\bf 0.039} & {\bf 0.034} & 0.041 & 0.054 & 0.049 & 0.050 & 0.086 & 0.089 & 0.179 & 0.039\\
NonInvasiveFetalECGThorax1 & {\bf 0.045} & {\bf 0.048} & {\bf 0.046} & {\bf 0.072} & 0.084 & 0.088 & 0.114 & 0.168 & 0.146 & 0.210 & 0.109\\
NonInvasiveFetalECGThorax2 & {\bf 0.046} & {\bf 0.050} & {\bf 0.044} & 0.057 & 0.069 & 0.075 & 0.107 & 0.171 & 0.135 & 0.205 & 0.105\\
OSULeaf & {\bf 0.052} & 0.060 & 0.057 & 0.123 & 0.085 & 0.086 & 0.123 & 0.119 & 0.132 & 0.193 & 0.084\\
OliveOil & {\bf 0.032} & 0.043 & {\bf 0.035} & 0.053 & 0.055 & 0.056 & 0.118 & 0.113 & 0.124 & 0.180 & 0.072\\
Phoneme & 0.401 & 0.433 & {\bf 0.382} & 0.468 & 0.646 & 0.442 & 0.496 & 0.510 & 0.484 & 0.581 & 0.452\\
PigAirwayPressure & 0.025 & 0.038 & 0.032 & 0.071 & 0.098 & 0.070 & 0.067 & 0.057 & 0.076 & 0.196 & {\bf 0.019}\\
PigArtPressure & {\bf 0.031} & 0.043 & 0.033 & 0.079 & 0.047 & 0.056 & 0.100 & 0.067 & 0.096 & 0.173 & 0.045\\
PigCVP & 0.078 & 0.084 & 0.076 & 0.101 & 0.077 & 0.084 & 0.082 & 0.084 & 0.105 & 0.174 & {\bf 0.070}\\
Plane & {\bf 0.105} & {\bf 0.111} & {\bf 0.105} & 0.160 & 0.138 & 0.135 & 0.200 & 0.294 & 0.245 & 0.316 & 0.189\\
PowerCons & 0.339 & {\bf 0.319} & 0.346 & 0.404 & 0.360 & 0.378 & 0.405 & 0.447 & 0.396 & 0.461 & 0.352\\
RefrigerationDevices & {\bf 0.577} & {\bf 0.579} & 0.589 & 0.606 & 0.591 & 0.648 & 0.612 & 0.593 & 0.613 & 0.656 & 0.606\\
Rock & 0.048 & 0.086 & 0.022 & 0.804 & 0.844 & 0.051 & 0.641 & 0.054 & 0.068 & 0.243 & {\bf 0.010}\\
ScreenType & {\bf 0.281} & 0.352 & {\bf 0.284} & 0.422 & {\bf 0.291} & {\bf 0.288} & 0.312 & 0.294 & {\bf 0.284} & 0.336 & 0.304\\
SemgHandGenderCh2 & {\bf 0.852} & 0.870 & 0.847 & {\bf 0.834} & 0.950 & 0.959 & 0.887 & 0.919 & 0.946 & 0.899 & 1.100\\
SemgHandMovementCh2 & {\bf 0.911} & {\bf 0.914} & {\bf 0.910} & {\bf 0.909} & 0.964 & 0.993 & 0.955 & 0.969 & 1.003 & 0.945 & 1.161\\
SemgHandSubjectCh2 & {\bf 0.826} & {\bf 0.833} & {\bf 0.825} & {\bf 0.827} & 0.900 & 0.934 & 0.859 & 0.898 & 0.924 & 0.876 & 1.072\\
ShapeletSim & 1.027 & {\bf 1.019} & 1.052 & {\bf 1.003} & 1.070 & 1.181 & 1.015 & 1.148 & 1.202 & 1.004 & 1.382\\
ShapesAll & {\bf 0.032} & 0.044 & {\bf 0.035} & 0.078 & 0.069 & 0.050 & 0.091 & 0.079 & 0.081 & 0.191 & 0.039\\
\hline
\end{tabular}}
  \end{scriptsize}
  \caption{Rooted mean squared error for each target task. Values in bold typeface are not statistically significantly different at the 5\% level from the best performing method in each row according to a paired t-test.}
  \label{tab:rmse1}
\end{table}

\begin{table}[t!]
  \centering
  \begin{scriptsize}
  {\tabcolsep=0.7em
    \begin{tabular}{l|r|rrr|rrr|rrr|r} \hline
      & & \multicolumn{3}{|c|}{LSTM} & \multicolumn{3}{|c|}{NN} & \multicolumn{3}{|c|}{Linear} & \\
      & Ours & MAML & DI & DS & MAML & DI & DS & MAML & DI & DS & Pre \\
      \hline
SmallKitchenAppliances & 1.082 & {\bf 1.077} & {\bf 1.079} & 1.086 & 1.162 & 1.096 & 1.111 & 1.271 & 1.243 & 1.207 & 1.277\\
StarLightCurves & 0.027 & 0.052 & 0.024 & 0.256 & 0.094 & 0.036 & 0.068 & 0.038 & 0.072 & 0.464 & {\bf 0.012}\\
Strawberry & {\bf 0.065} & {\bf 0.064} & 0.074 & 0.134 & 0.088 & 0.086 & 0.140 & 0.154 & 0.161 & 0.234 & 0.116\\
SwedishLeaf & 0.132 & {\bf 0.132} & {\bf 0.129} & 0.207 & 0.154 & 0.159 & 0.220 & 0.259 & 0.241 & 0.314 & 0.187\\
Symbols & {\bf 0.029} & 0.047 & {\bf 0.031} & 0.072 & 0.066 & 0.050 & 0.083 & 0.065 & 0.088 & 0.192 & 0.039\\
ToeSegmentation1 & {\bf 0.150} & 0.156 & {\bf 0.152} & {\bf 0.197} & {\bf 0.185} & {\bf 0.186} & 0.222 & 0.255 & 0.250 & 0.318 & 0.186\\
ToeSegmentation2 & {\bf 0.146} & 0.157 & 0.149 & {\bf 0.192} & {\bf 0.169} & {\bf 0.174} & 0.190 & 0.200 & 0.203 & 0.270 & 0.166\\
Trace & 0.123 & {\bf 0.114} & 0.124 & 0.160 & 0.160 & 0.164 & 0.195 & 0.227 & 0.210 & 0.285 & 0.162\\
TwoPatterns & 0.603 & {\bf 0.577} & 0.609 & {\bf 0.578} & 0.616 & 0.626 & {\bf 0.578} & 0.602 & 0.590 & 0.625 & 0.600\\
UMD & {\bf 0.078} & {\bf 0.080} & {\bf 0.076} & 0.153 & 0.132 & 0.113 & 0.133 & 0.151 & 0.155 & 0.261 & 0.104\\
UWaveGestureLibraryAll & {\bf 0.085} & 0.097 & 0.089 & 0.134 & 0.137 & {\bf 0.083} & 0.117 & 0.112 & 0.123 & 0.433 & {\bf 0.084}\\
UWaveGestureLibraryX & 0.041 & {\bf 0.064} & {\bf 0.036} & 0.109 & 0.114 & 0.067 & 0.094 & 0.100 & 0.110 & 0.423 & 0.053\\
UWaveGestureLibraryY & {\bf 0.030} & 0.060 & {\bf 0.029} & 0.088 & 0.040 & 0.036 & 0.123 & 0.067 & 0.093 & 0.197 & 0.043\\
UWaveGestureLibraryZ & {\bf 0.030} & 0.047 & {\bf 0.034} & 0.074 & 0.058 & 0.048 & 0.085 & 0.074 & 0.094 & 0.215 & 0.048\\
Wafer & {\bf 0.424} & 0.427 & 0.450 & {\bf 0.419} & 0.437 & 0.440 & 0.438 & 0.497 & 0.469 & 0.511 & 0.491\\
Wine & {\bf 0.126} & {\bf 0.112} & 0.121 & 0.233 & 0.198 & 0.215 & 0.256 & 0.334 & 0.302 & 0.353 & 0.227\\
WordSynonyms & {\bf 0.075} & {\bf 0.091} & 0.078 & 0.211 & 0.129 & 0.116 & 0.173 & 0.184 & 0.190 & 0.281 & 0.136\\
Worms & {\bf 0.068} & 0.086 & {\bf 0.068} & 0.117 & 0.091 & 0.075 & 0.108 & 0.098 & 0.107 & 0.196 & 0.072\\
WormsTwoClass & {\bf 0.067} & 0.095 & {\bf 0.068} & 0.119 & 0.089 & 0.083 & 0.124 & 0.090 & 0.110 & 0.196 & 0.072\\
Yoga & {\bf 0.032} & 0.041 & 0.037 & 0.111 & 0.053 & 0.057 & 0.089 & 0.083 & 0.101 & 0.176 & 0.055\\
\hline
Average & {\bf 0.224} &  0.235 &  0.231 &  0.295 &  0.293 &  0.272 &  0.299 &  0.305 &  0.312 &  0.387 &  0.285 \\
\hline
\#Best & {\bf 62} &  43 &  39 &  15 &  5 &  6 &  4 &  1 &  2 &  1 &  17 \\
\hline
\end{tabular}}
  \end{scriptsize}
  \caption{Rooted mean squared error for each target task continued from Table~\ref{tab:rmse1}. The second last row shows the avaraged rooted mean squared error over all target tasks. The last row shows the number of target datasets the method achieved the performance that are not statistically significantly different.}
  \label{tab:rmse2}
\end{table}

\begin{figure}[t!]
  \centering
  \includegraphics[width=20em]{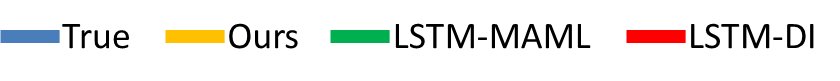}
  {\tabcolsep=-0.5em
  \begin{tabular}{ccc}
    \includegraphics[width=14em]{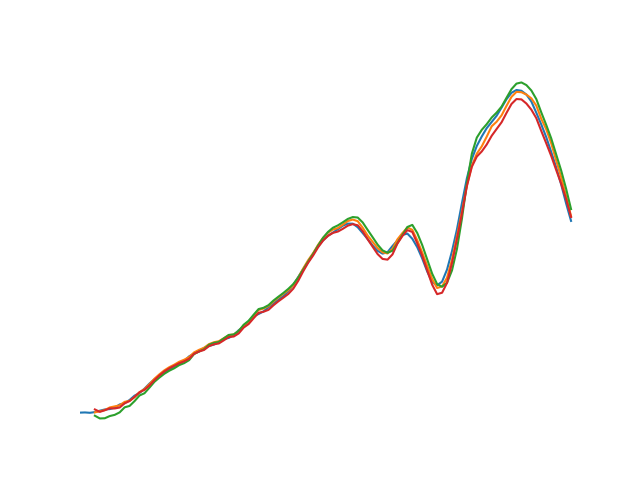}
    &
    \includegraphics[width=14em]{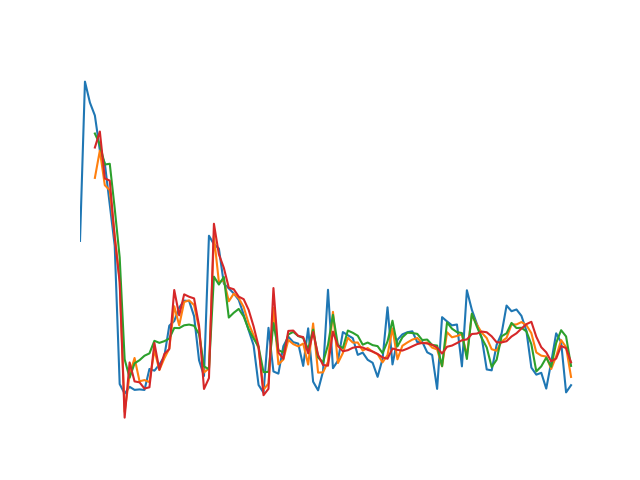}
    &
    \includegraphics[width=14em]{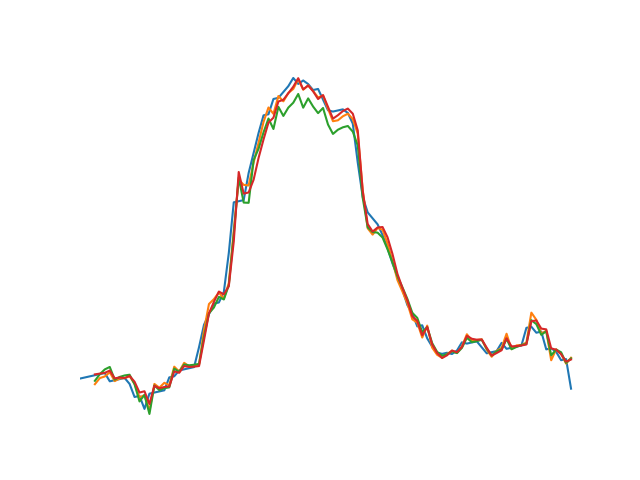}
    \\
    (a) Beef & (b) Chlorine & (c) EOG \\
    \includegraphics[width=14em]{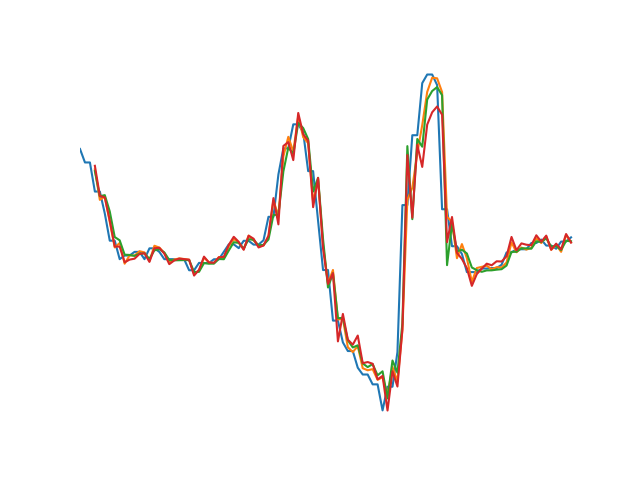}
    &
    \includegraphics[width=14em]{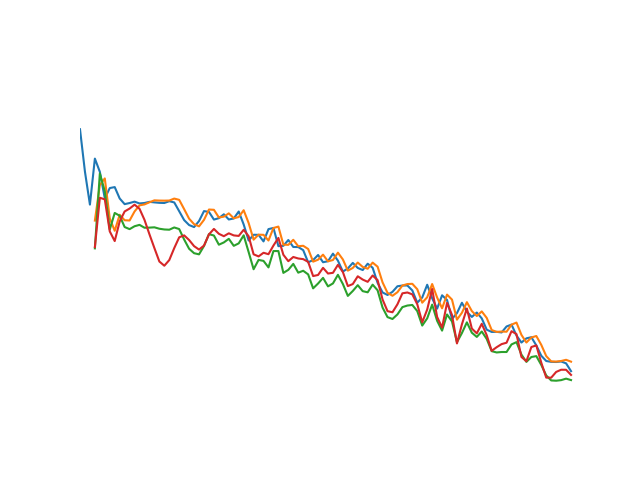}
    &
    \includegraphics[width=14em]{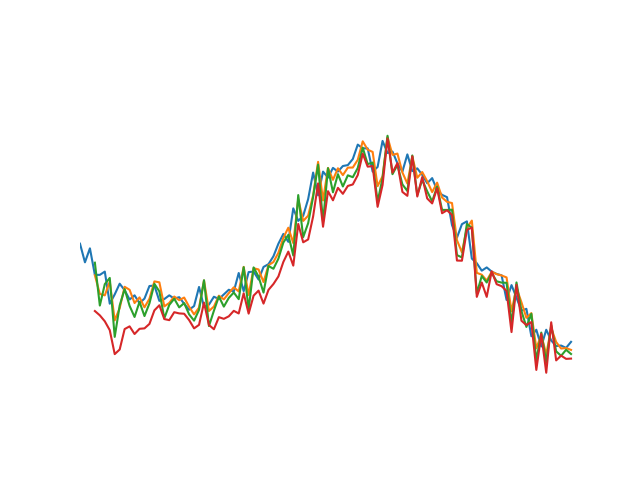}
    \\
    (d) FaceFour & (e) InlineSkate & (f) PigCVP \\
    \includegraphics[width=14em]{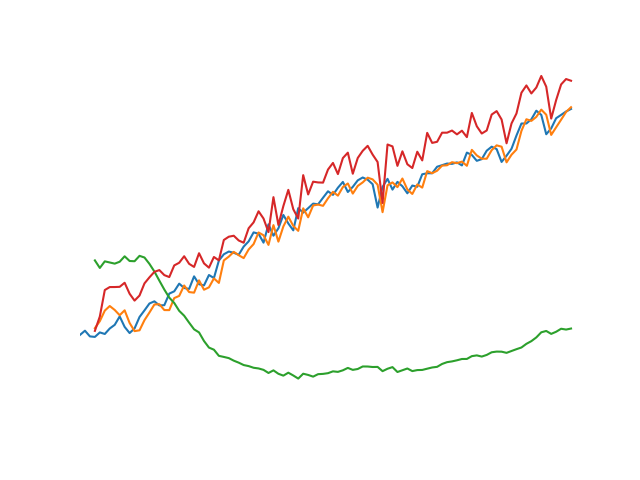}
    &
    \includegraphics[width=14em]{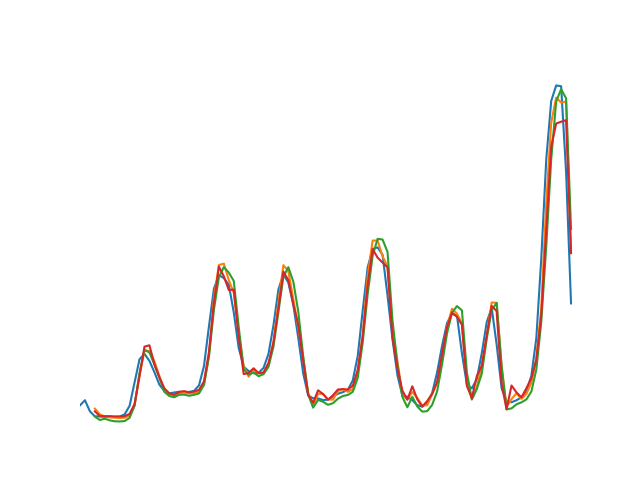}
    &
    \includegraphics[width=14em]{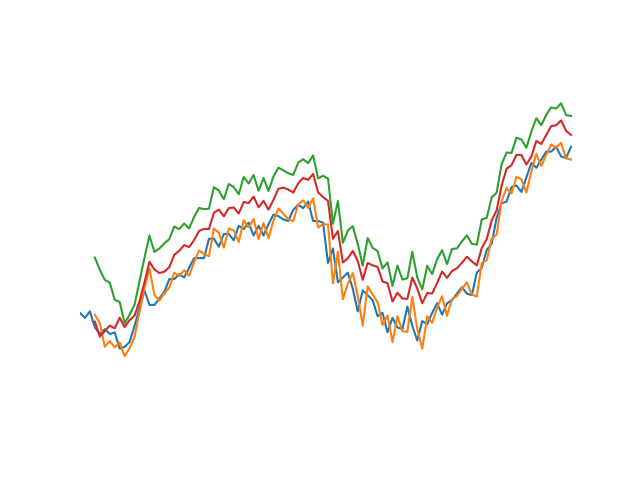}\\
    (g) Rock & (h) WordSynonyms & (i) Worms \\
  \end{tabular}}
  \caption{Examples of true time-series values (blue), and next timestep forecasting results with the proposed method (yellow), LSTM-MAML (green), and LSTM-DI (red).}
  \label{fig:time-series}
\end{figure}

Tables~\ref{tab:rmse1} and \ref{tab:rmse2} show the rooted mean squared error of next timestep forecasting
for each target task averaged over 30 experiments with different training, validation and target splits.
The proposed method achieved the performance that was not different from the best method
62 among 90 target tasks, which was the most among comparing methods.
Generally, LSTM was better than NN, and NN was better than Linear.
This result indicates that LSTM-based recurrent neural networks are appropriate for forecasting time-series.
LSTM-MAML was worse than the proposed method.
The reason is that time-series dynamics are very different across tasks,
and it is difficult to finetune well from a single initial model parameter setting
for diverse tasks.
On the other hand, the proposed method flexibly adapt to target tasks with attention mechanisms
given the support set.
LSTM-DI performed similar performance to LSTM-MAML.
Although LSTM-DI does not use support sets of target tasks,
it can give task-specific forecasting by taking query time-series as input with LSTM.
Figure~\ref{fig:time-series} shows some examples of true and forecasted values by the proposed method, LSTM-MAML, and LSTM-DI.
The proposed method forecasted appropriately with different dynamics of target tasks.

\begin{figure}[t!]
  \centering
  {\tabcolsep=0em
  \begin{tabular}{cc}
    \includegraphics[width=20em]{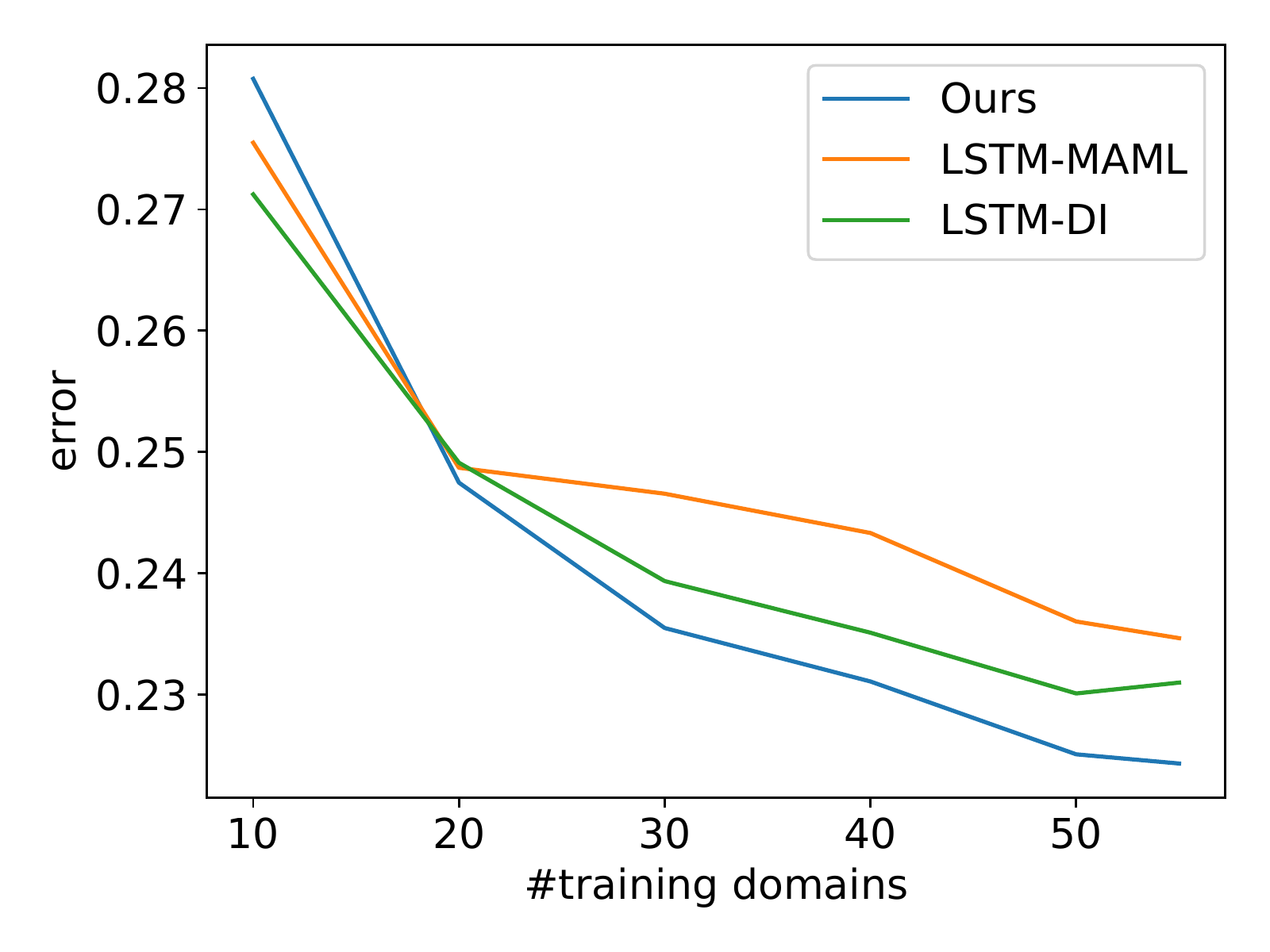} &
    \includegraphics[width=20em]{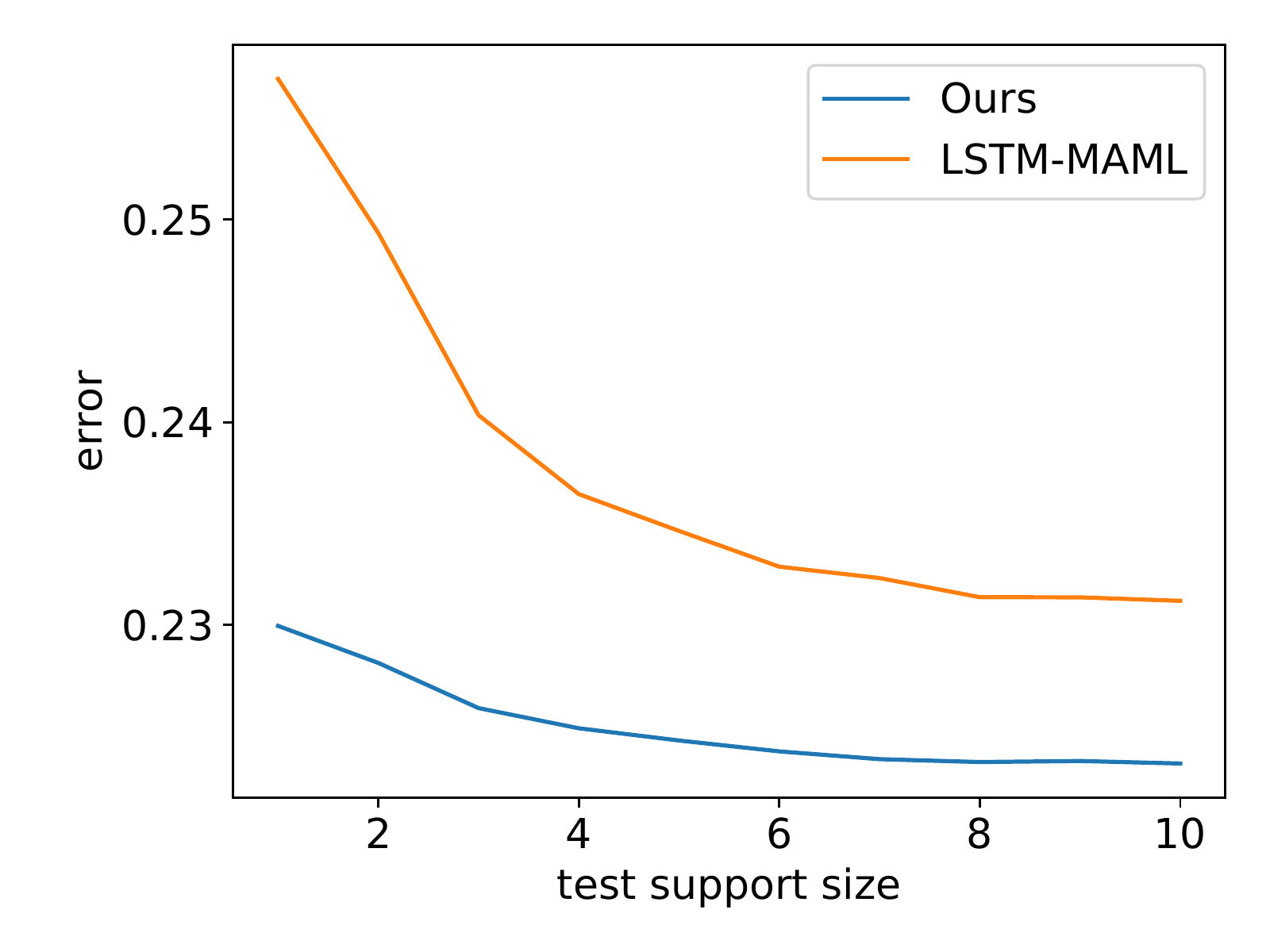} \\
    (a) \#training tasks &
    (b) Test support size\\
  \end{tabular}}
  \caption{Rooted mean squared error averaged over all target tasks (a) with different numbers of training tasks, and (b) with different test suport size.}
  \label{fig:err}
\end{figure}

Figure~\ref{fig:err}(a) shows the average mean squared error with different numbers of training tasks
by the proposed method, LSTM-MAML, and LSTM-DI. All the methods decreased the errors as the number of training tasks increased.
Figure~\ref{fig:err}(b) shows the average mean squared error with different test support size
by the proposed method and LSTM-MAML, where the training support size was three.
Even when the test support size was different from training,
the proposed method and LSTM-MAML decreased the error as the test support size increased.
Table~\ref{tab:time} shows
the average computational time in seconds of training with all training tasks, and computational time of test for each target task.
on computers with 2.30GHz CPUs with five cores.
The proposed method had slightly shorter training and test time than LSTM-MAML.

\begin{table}[t!]
  \centering
  \caption{Average computational time in seconds of training with all training tasks, and computational time of test for each target task.}
  \label{tab:time}
  \begin{tabular}{lrrr}
    \hline
    & Ours & LSTM-MAML & LSTM-DI\\
    \hline
    Train & 40,203 & 47,978 & 5,297 \\
    Test & 15 & 20 & 2 \\
    \hline
  \end{tabular}
\end{table}

\section{Conclusion}
\label{sec:conclusion}

In this paper, we proposed a meta-learning method for time-series forecasting,
where our model is trained with many time-series datasets.
Our model can forecast future values that are specific to a target task
using a few time-series data in the target task by recurrent neural networks with an attention mechanism.
For future work, we plan to apply the proposed method to multivariate time-series datasets.

\bibliography{neurips2020}

\begin{thebibliography}{10}

\bibitem{ali2018cross}
A.~R. Ali, B.~Gabrys, and M.~Budka.
\newblock Cross-domain meta-learning for time-series forecasting.
\newblock {\em Procedia Computer Science}, 126:9--18, 2018.

\bibitem{andrychowicz2016learning}
M.~Andrychowicz, M.~Denil, S.~Gomez, M.~W. Hoffman, D.~Pfau, T.~Schaul,
  B.~Shillingford, and N.~De~Freitas.
\newblock Learning to learn by gradient descent by gradient descent.
\newblock In {\em Advances in Neural Information Processing Systems}, pages
  3981--3989, 2016.

\bibitem{assaad2008new}
M.~Assaad, R.~Bon{\'e}, and H.~Cardot.
\newblock A new boosting algorithm for improved time-series forecasting with
  recurrent neural networks.
\newblock {\em Information Fusion}, 9(1):41--55, 2008.

\bibitem{azoff1994neural}
E.~M. Azoff.
\newblock {\em Neural Network Time Series Forecasting of Financial Markets}.
\newblock John Wiley \& Sons, Inc., 1994.

\bibitem{bartunov2018few}
S.~Bartunov and D.~Vetrov.
\newblock Few-shot generative modelling with generative matching networks.
\newblock In {\em International Conference on Artificial Intelligence and
  Statistics}, pages 670--678, 2018.

\bibitem{bengio1991learning}
Y.~Bengio, S.~Bengio, and J.~Cloutier.
\newblock Learning a synaptic learning rule.
\newblock In {\em International Joint Conference on Neural Networks}, 1991.

\bibitem{bornschein2017variational}
J.~Bornschein, A.~Mnih, D.~Zoran, and D.~J. Rezende.
\newblock Variational memory addressing in generative models.
\newblock In {\em Advances in Neural Information Processing Systems}, pages
  3920--3929, 2017.

\bibitem{cao2003support}
L.-J. Cao and F.~E.~H. Tay.
\newblock Support vector machine with adaptive parameters in financial time
  series forecasting.
\newblock {\em IEEE Transactions on Neural Networks}, 14(6):1506--1518, 2003.

\bibitem{dau2019ucr}
H.~A. Dau, A.~Bagnall, K.~Kamgar, C.-C.~M. Yeh, Y.~Zhu, S.~Gharghabi, C.~A.
  Ratanamahatana, and E.~Keogh.
\newblock The {UCR} time series archive.
\newblock {\em IEEE/CAA Journal of Automatica Sinica}, 6(6):1293--1305, 2019.

\bibitem{UCRArchive2018}
H.~A. Dau, E.~Keogh, K.~Kamgar, C.-C.~M. Yeh, Y.~Zhu, S.~Gharghabi, C.~A.
  Ratanamahatana, Yanping, B.~Hu, N.~Begum, A.~Bagnall, A.~Mueen, G.~Batista,
  and Hexagon-ML.
\newblock The {UCR} time series classification archive, October 2018.
\newblock \url{https://www.cs.ucr.edu/~eamonn/time_series_data_2018/}.

\bibitem{deb2017review}
C.~Deb, F.~Zhang, J.~Yang, S.~E. Lee, and K.~W. Shah.
\newblock A review on time series forecasting techniques for building energy
  consumption.
\newblock {\em Renewable and Sustainable Energy Reviews}, 74:902--924, 2017.

\bibitem{edwards2016towards}
H.~Edwards and A.~Storkey.
\newblock Towards a neural statistician.
\newblock {\em arXiv preprint arXiv:1606.02185}, 2016.

\bibitem{finn2017model}
C.~Finn, P.~Abbeel, and S.~Levine.
\newblock Model-agnostic meta-learning for fast adaptation of deep networks.
\newblock In {\em Proceedings of the 34th International Conference on Machine
  Learning}, pages 1126--1135, 2017.

\bibitem{finn2018probabilistic}
C.~Finn, K.~Xu, and S.~Levine.
\newblock Probabilistic model-agnostic meta-learning.
\newblock In {\em Advances in Neural Information Processing Systems}, pages
  9516--9527, 2018.

\bibitem{garnelo2018conditional}
M.~Garnelo, D.~Rosenbaum, C.~Maddison, T.~Ramalho, D.~Saxton, M.~Shanahan,
  Y.~W. Teh, D.~Rezende, and S.~A. Eslami.
\newblock Conditional neural processes.
\newblock In {\em International Conference on Machine Learning}, pages
  1690--1699, 2018.

\bibitem{hewitt2018variational}
L.~B. Hewitt, M.~I. Nye, A.~Gane, T.~Jaakkola, and J.~B. Tenenbaum.
\newblock The variational homoencoder: Learning to learn high capacity
  generative models from few examples.
\newblock {\em arXiv preprint arXiv:1807.08919}, 2018.

\bibitem{hochreiter1997long}
S.~Hochreiter and J.~Schmidhuber.
\newblock Long short-term memory.
\newblock {\em Neural Computation}, 9(8):1735--1780, 1997.

\bibitem{hooshmand2019energy}
A.~Hooshmand and R.~Sharma.
\newblock Energy predictive models with limited data using transfer learning.
\newblock In {\em Proceedings of the Tenth ACM International Conference on
  Future Energy Systems}, pages 12--16, 2019.

\bibitem{jia2018transfer}
Y.~Jia, Y.~Zhang, R.~Weiss, Q.~Wang, J.~Shen, F.~Ren, P.~Nguyen, R.~Pang, I.~L.
  Moreno, Y.~Wu, et~al.
\newblock Transfer learning from speaker verification to multispeaker
  text-to-speech synthesis.
\newblock In {\em Advances in Neural Information Processing Systems}, pages
  4480--4490, 2018.

\bibitem{jilani2007multivariate}
T.~A. Jilani, S.~A. Burney, and C.~Ardil.
\newblock Multivariate high order fuzzy time series forecasting for car road
  accidents.
\newblock {\em International Journal of Computational Intelligence},
  4(1):15--20, 2007.

\bibitem{killian2017robust}
T.~W. Killian, S.~Daulton, G.~Konidaris, and F.~Doshi-Velez.
\newblock Robust and efficient transfer learning with hidden parameter markov
  decision processes.
\newblock In {\em Advances in Neural Information Processing Systems}, pages
  6250--6261, 2017.

\bibitem{kim2019attentive}
H.~Kim, A.~Mnih, J.~Schwarz, M.~Garnelo, A.~Eslami, D.~Rosenbaum, O.~Vinyals,
  and Y.~W. Teh.
\newblock Attentive neural processes.
\newblock {\em arXiv preprint arXiv:1901.05761}, 2019.

\bibitem{kim2003financial}
K.-j. Kim.
\newblock Financial time series forecasting using support vector machines.
\newblock {\em Neurocomputing}, 55(1-2):307--319, 2003.

\bibitem{kimbayesian}
T.~Kim, J.~Yoon, O.~Dia, S.~Kim, Y.~Bengio, and S.~Ahn.
\newblock Bayesian model-agnostic meta-learning.
\newblock In {\em Advances in Neural Information Processing Systems}, 2018.

\bibitem{kingma2014adam}
D.~P. Kingma and J.~Ba.
\newblock {ADAM}: {A} method for stochastic optimization.
\newblock In {\em International Conference on Learning Representations}, 2015.

\bibitem{kumagai2019transfer}
A.~Kumagai, T.~Iwata, and Y.~Fujiwara.
\newblock Transfer anomaly detection by inferring latent domain
  representations.
\newblock In {\em Advances in Neural Information Processing Systems}, pages
  2467--2477, 2019.

\bibitem{lachtermacher1994backpropagation}
G.~Lachtermacher and J.~D. Fuller.
\newblock Backpropagation in hydrological time series forecasting.
\newblock In {\em Stochastic and Statistical Methods in Hydrology and
  Environmental Engineering}, pages 229--242. Springer, 1994.

\bibitem{lake2019compositional}
B.~M. Lake.
\newblock Compositional generalization through meta sequence-to-sequence
  learning.
\newblock In {\em Advances in Neural Information Processing Systems}, pages
  9788--9798, 2019.

\bibitem{laptev2017time}
N.~Laptev, J.~Yosinski, L.~E. Li, and S.~Smyl.
\newblock Time-series extreme event forecasting with neural networks at uber.
\newblock In {\em International Conference on Machine Learning}, volume~34,
  pages 1--5, 2017.

\bibitem{lemke2010meta}
C.~Lemke and B.~Gabrys.
\newblock Meta-learning for time series forecasting and forecast combination.
\newblock {\em Neurocomputing}, 73(10-12):2006--2016, 2010.

\bibitem{li2017diffusion}
Y.~Li, R.~Yu, C.~Shahabi, and Y.~Liu.
\newblock Diffusion convolutional recurrent neural network: Data-driven traffic
  forecasting.
\newblock {\em arXiv preprint arXiv:1707.01926}, 2017.

\bibitem{li2017meta}
Z.~Li, F.~Zhou, F.~Chen, and H.~Li.
\newblock Meta-{SGD}: Learning to learn quickly for few-shot learning.
\newblock {\em arXiv preprint arXiv:1707.09835}, 2017.

\bibitem{long2017deep}
M.~Long, H.~Zhu, J.~Wang, and M.~I. Jordan.
\newblock Deep transfer learning with joint adaptation networks.
\newblock In {\em Proceedings of the 34th International Conference on Machine
  Learning}, pages 2208--2217, 2017.

\bibitem{narwariya2020meta}
J.~Narwariya, P.~Malhotra, L.~Vig, G.~Shroff, and T.~Vishnu.
\newblock Meta-learning for few-shot time series classification.
\newblock In {\em Proceedings of the 7th ACM IKDD CoDS and 25th COMAD}, pages
  28--36. 2020.

\bibitem{ogunmolu2016nonlinear}
O.~Ogunmolu, X.~Gu, S.~Jiang, and N.~Gans.
\newblock Nonlinear systems identification using deep dynamic neural networks.
\newblock {\em arXiv preprint arXiv:1610.01439}, 2016.

\bibitem{oreshkin2019n}
B.~N. Oreshkin, D.~Carpov, N.~Chapados, and Y.~Bengio.
\newblock {N-BEATS}: Neural basis expansion analysis for interpretable time
  series forecasting.
\newblock {\em arXiv preprint arXiv:1905.10437}, 2019.

\bibitem{oreshkin2020meta}
B.~N. Oreshkin, D.~Carpov, N.~Chapados, and Y.~Bengio.
\newblock Meta-learning framework with applications to zero-shot time-series
  forecasting.
\newblock {\em arXiv preprint arXiv:2002.02887}, 2020.

\bibitem{prudencio2004meta}
R.~B. Prud{\^e}ncio and T.~B. Ludermir.
\newblock Meta-learning approaches to selecting time series models.
\newblock {\em Neurocomputing}, 61:121--137, 2004.

\bibitem{qin2019recurrent}
S.~Qin, J.~Zhu, J.~Qin, W.~Wang, and D.~Zhao.
\newblock Recurrent attentive neural process for sequential data.
\newblock {\em arXiv preprint arXiv:1910.09323}, 2019.

\bibitem{ravi2016optimization}
S.~Ravi and H.~Larochelle.
\newblock Optimization as a model for few-shot learning.
\newblock In {\em International Conference on Learning Representations}, 2017.

\bibitem{reed2017few}
S.~Reed, Y.~Chen, T.~Paine, A.~v.~d. Oord, S.~Eslami, D.~Rezende, O.~Vinyals,
  and N.~de~Freitas.
\newblock Few-shot autoregressive density estimation: Towards learning to learn
  distributions.
\newblock {\em arXiv preprint arXiv:1710.10304}, 2017.

\bibitem{rezende2016one}
D.~J. Rezende, S.~Mohamed, I.~Danihelka, K.~Gregor, and D.~Wierstra.
\newblock One-shot generalization in deep generative models.
\newblock In {\em Proceedings of the 33rd International Conference on
  International Conference on Machine Learning}, pages 1521--1529, 2016.

\bibitem{ribeiro2018transfer}
M.~Ribeiro, K.~Grolinger, H.~F. ElYamany, W.~A. Higashino, and M.~A. Capretz.
\newblock Transfer learning with seasonal and trend adjustment for
  cross-building energy forecasting.
\newblock {\em Energy and Buildings}, 165:352--363, 2018.

\bibitem{rusu2018meta}
A.~A. Rusu, D.~Rao, J.~Sygnowski, O.~Vinyals, R.~Pascanu, S.~Osindero, and
  R.~Hadsell.
\newblock Meta-learning with latent embedding optimization.
\newblock In {\em International Conference on Learning Representations}, 2019.

\bibitem{schmidhuber:1987:srl}
J.~Schmidhuber.
\newblock Evolutionary principles in self-referential learning. on learning now
  to learn: The meta-meta-meta...-hook.
\newblock Master's thesis, Technische Universitat Munchen, Germany, 1987.

\bibitem{sfetsos2000comparison}
A.~Sfetsos.
\newblock A comparison of various forecasting techniques applied to mean hourly
  wind speed time series.
\newblock {\em Renewable Energy}, 21(1):23--35, 2000.

\bibitem{snell2017prototypical}
J.~Snell, K.~Swersky, and R.~Zemel.
\newblock Prototypical networks for few-shot learning.
\newblock In {\em Advances in Neural Information Processing Systems}, pages
  4077--4087, 2017.

\bibitem{talagala2018meta}
T.~S. Talagala, R.~J. Hyndman, G.~Athanasopoulos, et~al.
\newblock Meta-learning how to forecast time series.
\newblock {\em Monash Econometrics and Business Statistics Working Papers},
  6:18, 2018.

\bibitem{tan2018survey}
C.~Tan, F.~Sun, T.~Kong, W.~Zhang, C.~Yang, and C.~Liu.
\newblock A survey on deep transfer learning.
\newblock In {\em International Conference on Artificial Neural Networks},
  pages 270--279. Springer, 2018.

\bibitem{tang2019}
W.~Tang, L.~Liu, and G.~Long.
\newblock Few-shot time-series classification with dual interpretability.
\newblock In {\em ICML Time Series Workshop}. 2019.

\bibitem{vinyals2016matching}
O.~Vinyals, C.~Blundell, T.~Lillicrap, D.~Wierstra, et~al.
\newblock Matching networks for one shot learning.
\newblock In {\em Advances in neural information processing systems}, pages
  3630--3638, 2016.

\bibitem{willi2019recurrent}
T.~Willi, J.~Masci, J.~Schmidhuber, and C.~Osendorfer.
\newblock Recurrent neural processes.
\newblock {\em arXiv preprint arXiv:1906.05915}, 2019.

\bibitem{xie2019meta}
Y.~Xie, H.~Jiang, F.~Liu, T.~Zhao, and H.~Zha.
\newblock Meta learning with relational information for short sequences.
\newblock In {\em Advances in Neural Information Processing Systems}, pages
  9901--9912, 2019.

\bibitem{yao2019hierarchically}
H.~Yao, Y.~Wei, J.~Huang, and Z.~Li.
\newblock Hierarchically structured meta-learning.
\newblock In {\em International Conference on Machine Learning}, pages
  7045--7054, 2019.

\bibitem{yu2017spatio}
B.~Yu, H.~Yin, and Z.~Zhu.
\newblock Spatio-temporal graph convolutional neural network: A deep learning
  framework for traffic forecasting.

\end{thebibliography}
\bibliographystyle{abbrv}

\end{document}